# Place Recognition with Event-based Cameras and a Neural Implementation of SeqSLAM

Michael Milford[1], Hanme Kim[2], Michael Mangan[3], Tom Stone[3], Stefan Leutenegger[2], Barbara Webb[3] and Andrew Davison[2]

[1]Australian Centre for Robotic Vision, Queensland University of Technology
[2]Department of Computing, Imperial College
[3]School of Informatics, University of Edinburgh
Corresponding author: michael.milford@qut.edu.au

## 1      Abstract

Event-based cameras (Figure 1) offer much potential to the fields of robotics and computer vision, in part due to their large dynamic range and extremely high "frame rates". These attributes make them, at least in theory, particularly suitable for enabling tasks like navigation and mapping on high speed robotic platforms under challenging lighting conditions, a task which has been particularly challenging for traditional algorithms and camera sensors. Before these tasks become feasible however, progress must be made towards adapting and innovating current RGB-camera-based algorithms to work with event-based cameras. In this paper we present ongoing research investigating two distinct approaches to incorporating event-based cameras for robotic navigation:

1. The investigation of suitable place recognition / loop closure techniques, and
2. The development of efficient neural implementations of place recognition techniques that enable the possibility of place recognition using event-based cameras at very high frame rates using neuromorphic computing hardware.

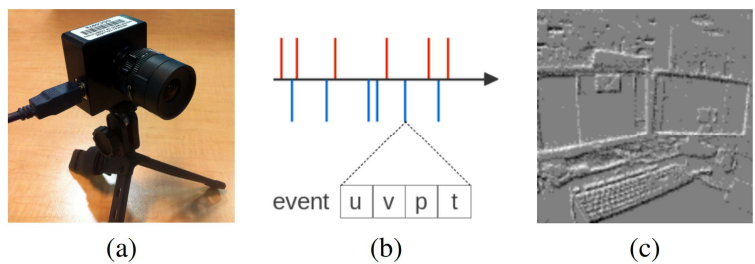

Figure 1: The first commercial event camera: (a) DVS128; (b) a stream of events (upward and downward spikes: positive and negative events); (c) image-like visualisation of accumulated events within a time interval (white and black: positive and negative events). From  (H. Kim, 2014)].

## 2      Place Recognition using SeqSLAM

SeqSLAM is a state-of-the-art algorithm for performing place recognition using camera frame sequences (Michael Milford, 2012). It uses low quality, low resolution imagery from RGB cameras under challenging lighting conditions, rendering it a potentially suitable algorithm for performing loop closure using the sparse, low resolution output from an event camera. For these initial experiments, we used the open source version OpenSeqSLAM, available from <http://openslam.org/>, using a sequence length of 100 frames. For these initial experiments, we assumed an approximate camera translational speed measure was available (such that might be obtained from a visual odometry system once implemented), in order to reduce the size of the search space for SeqSLAM.

We gathered three datasets at 3 different average speeds in an office environment from a forward facing DVS128 event camera being carried by an experimenter (Figure 2):

- **Run A:** running
- **Run B:** walking 66% of Run A speed
- **Run C:** slow walking 25% of Run A speed

Events were accumulated into 10 ms time windows (effective fps of 100) to form 128 × 128 event snapshots, which were downsampled to 16 × 16 pixel resolution then input into OpenSeqSLAM. We ran two experiments evaluating place recognition performance from the two slower traverses (Runs B-C) matching back to the fastest traverse (Run A). Approximate frame correspondence ground truth was obtained by manually inspecting the frames and is shown in Figure 2b. Each traverse contained internal loops.

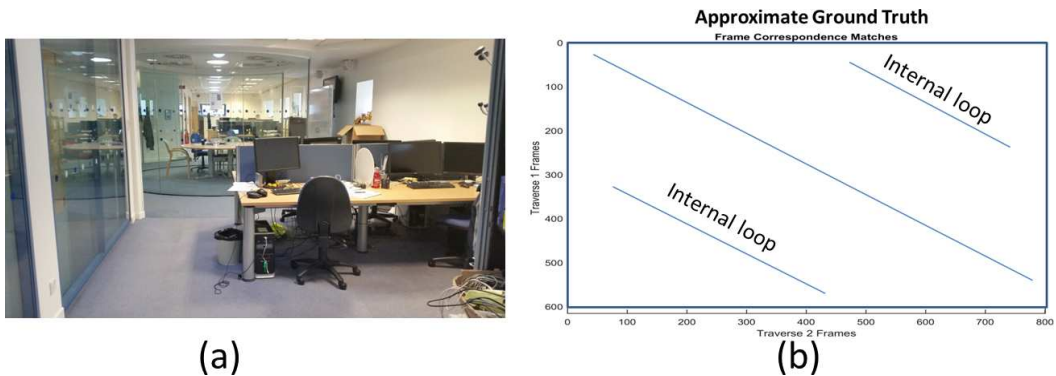

(a) (b)

Figure 2: (a) Office testing environment and (b) approximate frame correspondence ground truth for matches between two traverses, obtained by manual inspection of video frames. Note the internal loop closures within each dataset.

Figure 3 shows the frame correspondence matches between Runs A and B, overlaid on the SeqSLAM confusion matrix. Almost 100% match coverage is obtainable with no significant localization errors.

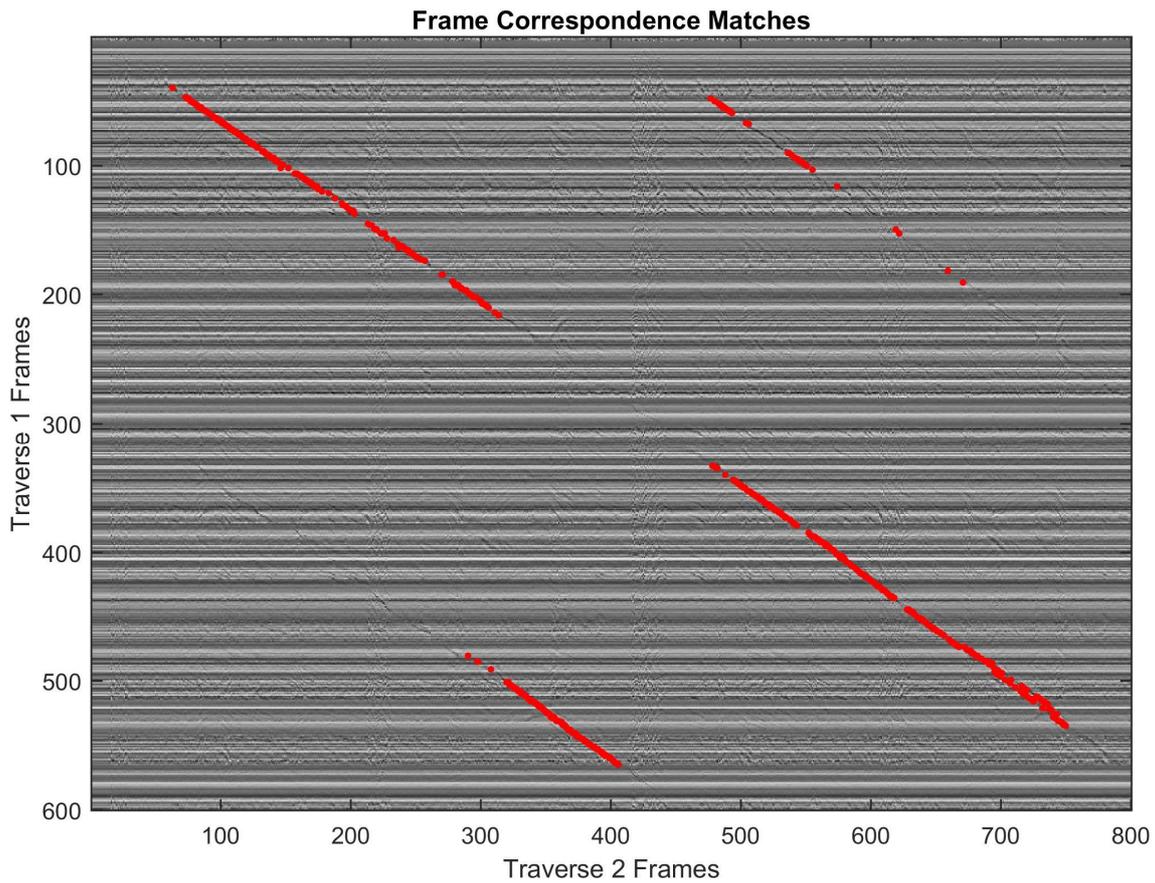

Figure 3: Frame matching for Runs A & B, with a 1.5 × speed differential.

Figure 4 shows the frame matches obtained between Runs A and C, for a speed differential of approximately 4 times. Matching coverage is reduced but still significant, with correct frame matches obtained over approximately 30% of the dataset, which would likely be sufficient for any SLAM system.

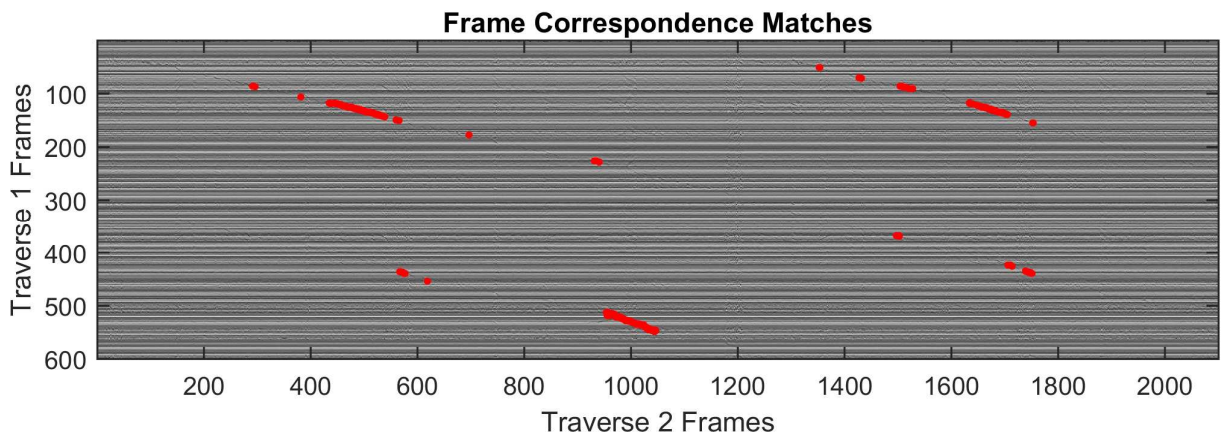

Figure 4: Frame matching for Runs A & C, with an approximately 4× speed differential.

Figure 5 shows sample frames from sequences that were successfully matched between Runs A and C. The matches were made despite large variations in camera speed and camera jitter that significantly changes both the overall event image and specific details such as polarity (for example polarity is partially switched in (b) due to opposite vertical camera jitter).

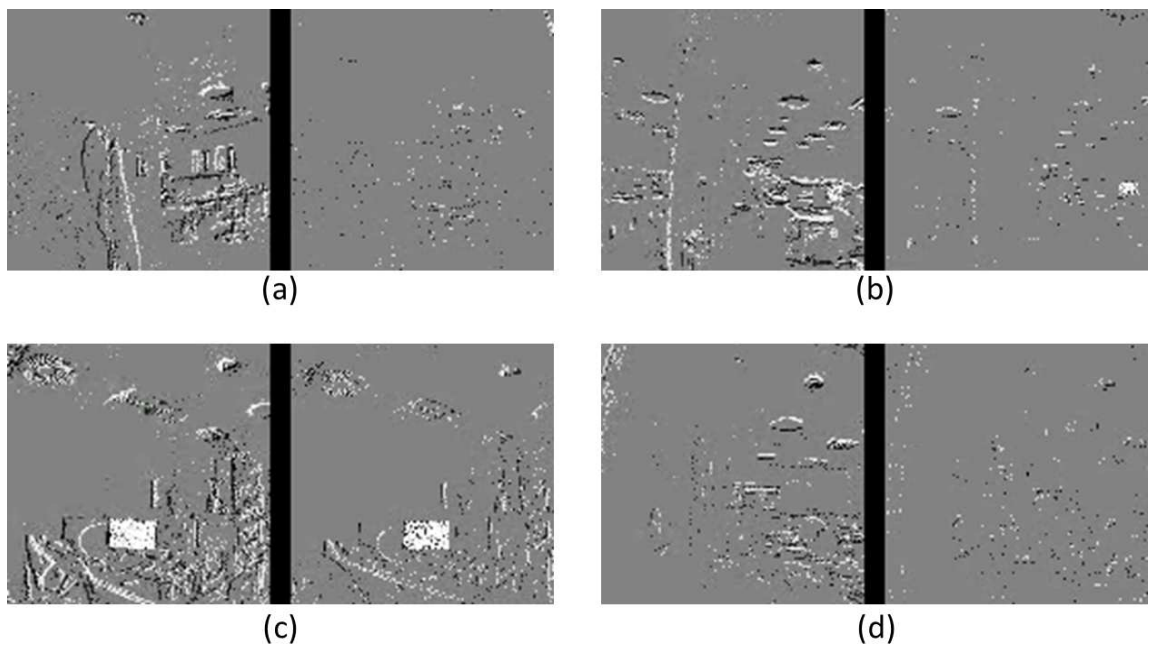

Figure 5: Sample frame matches between the fastest (A) and slowest (C) runs.

For future event camera-based SLAM implementations, it is clear that traditional feature-based techniques will require significant adaptation (or boosting through additional sensory modalities) in order to achieve robust place recognition performance, given the nature of the data. Here we have shown that a current technique, SeqSLAM, performs place recognition on event camera data despite significant changes in camera speed and hence significantly different event "snapshots".

## 3   Neural SeqSLAM

Current algorithmic implementations of SeqSLAM have two major disadvantages with regards to efficient implementation on low power computational hardware; images are stored as templates, with growing storage requirements as the environment size increases, and core computation involves operations that do not efficiently translate to neuromorphic computational hardware. Especially interesting is the potential for implementing robotic algorithms on spiking hardware (Francesco Galluppi, 2012). From a biological perspective, it is also interesting to examine whether visual navigation algorithms like SeqSLAM can be implemented in a neural model that is plausible for small organisms such as ants (Thomas Stone, 2014).

Here we present a lightweight neural implementation of SeqSLAM and demonstrate that it can perform sequence-based place recognition in places that have undergone moderate appearance-change. Input images (downsampled to 32 × 24 pixels and patch normalized) are fed into the network's input layer (Figure 6); these images are then pattern separated through sparsification into a much larger layer, before being consolidated in the output layer. Recurrent connectivity in the output layer results in a permanent dynamic of sideways shifting activity, much like some of the postulated mechanisms for how rotation is encoded in the rodent brain. Recently there has been evidence of similar one-dimensional ring network attractors in the insect brain as well. To provide a sequence-like matching capability, the network learns this recurrent connectivity based on the activation order of the output neurons during the training traverse of the environment. Table 1 shows the network parameters for the experiments described here.

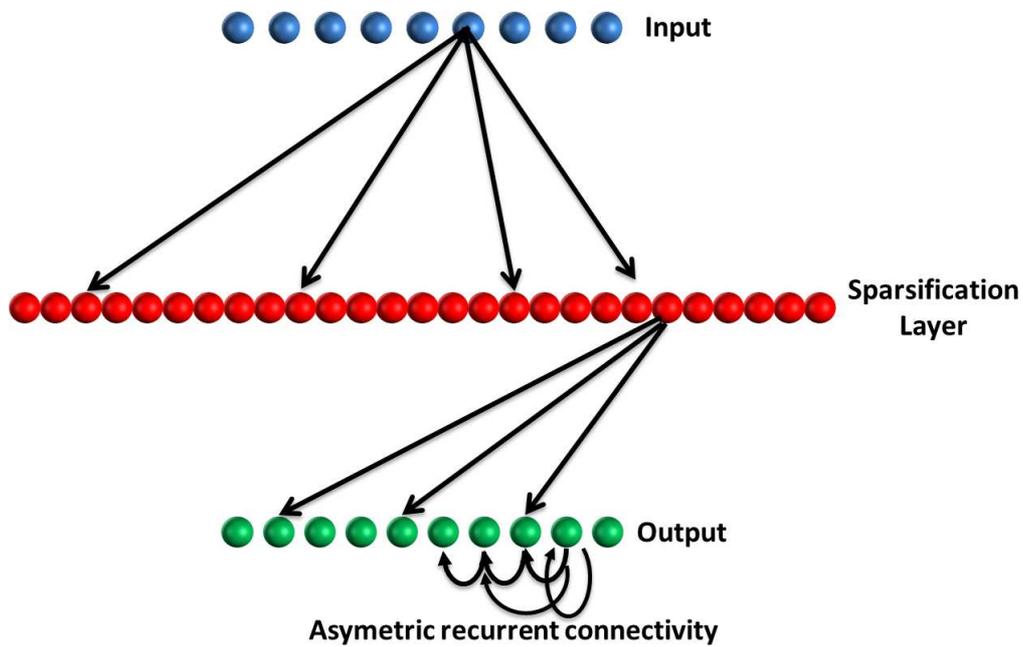

Figure 6: Network architecture.

Table 1: Network parameters.

| Model Component | Parameter Details |
|---|---|
| **Input Layer (L1)** | 768 units |
| **L1-L2 connectivity density** | 20% of full interconnectivity |
| **Sparsification Layer (L2)** | 3072 units |
| **L2-L3 connectivity density** | 20% of full interconnectivity |
| **Output Layer (L3)** | 555 units |

Figure 7 shows the activation rates of the output layer (columns in the matrix) during training and testing for two simulated environment traverses using synthetic data. Black crosses indicate the maximally firing output units, which can be interpreted as the current place match hypothesis. During training, the place match hypotheses are often incorrect, but during testing after the recurrency has been learnt, the system exhibits near perfect performance.

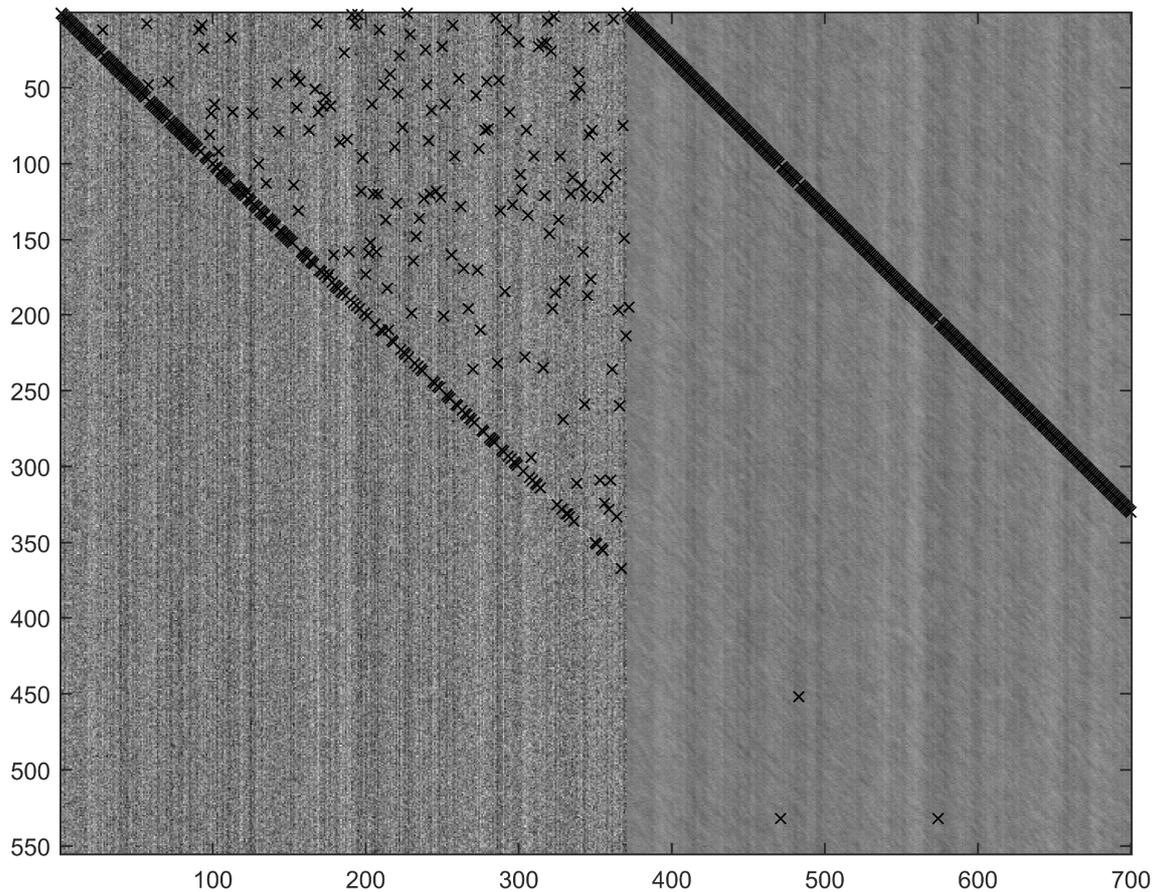

Figure 7: Sequence learning (first half) and recognition (second half) using two synthetic image data image traverses.

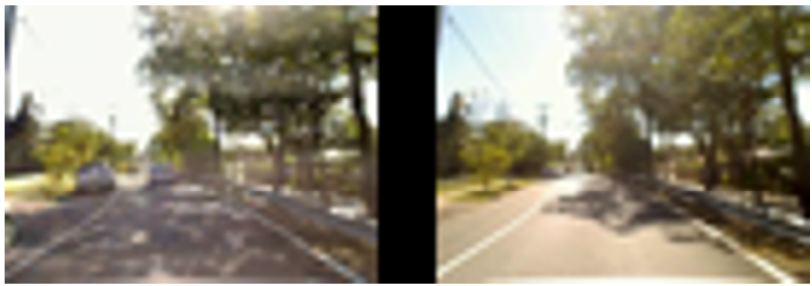

Figure 8: Sample frames from the sequence learning and recognition experiments using real world image data with moderate appearance change.

Figure 9 shows the activation rates of the output layer (columns in the matrix) during training and testing for two traverses through an environment with moderate appearance change (see Figure 8). Although place recognition performance is not as clean, the system after training is still able to correctly recognize about 80% of places along the traverse, despite the moderate appearance change. Figure 10 shows examples of pairs of corresponding frames from matched sequences.

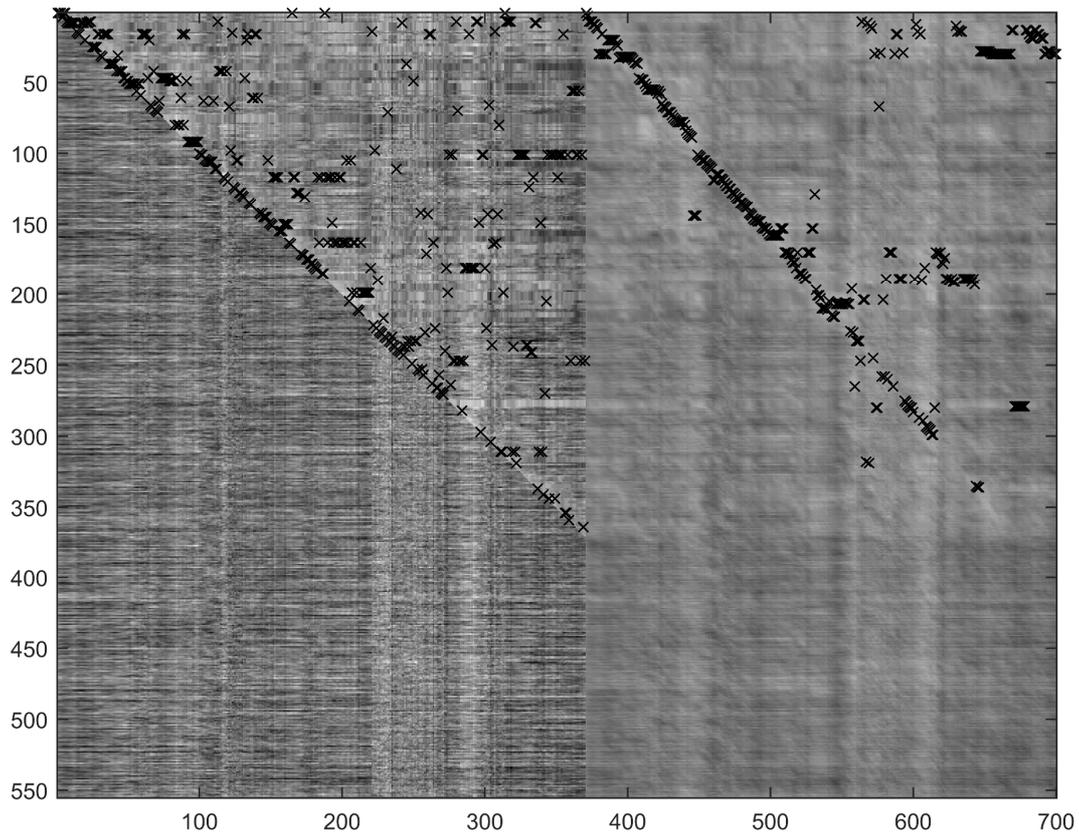

Figure 9: Sequence learning (first half) and recognition (second half) using real-world image data from an environment with moderate illumination change.

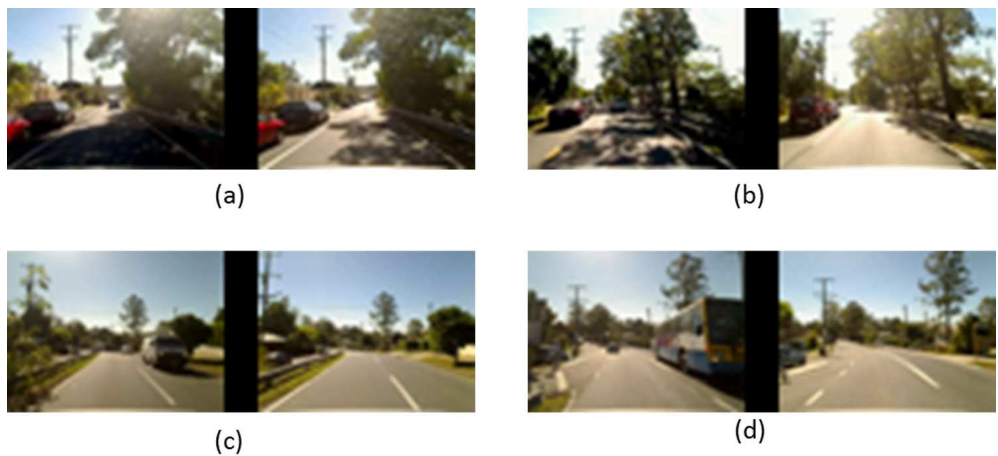

Figure 10: Sample frame matches between matched sequences from the two real-world traverses.

The current neural implementation has two major limitations. The first is that it cannot yet handle extreme appearance change like the conventional SeqSLAM algorithm can – this is partly because the network matching output does not degrade gracefully as the degree of appearance change increases; in comparison SeqSLAM explicitly stores all the original image templates from the training run for direct image comparison during testing, which leads to a more graceful degradation in matching performance. SeqSLAM also performs a form of cohort normalization, which we may be able to replicate by implementing multiple continuous attractor networks in parallel, each responsible for encoding a subsection of the environment.

Secondly, the naïve network structure does not deal with the highly repetitive structure of images along a route, especially in the demonstrated car-based navigation scenario – this is why the synthetic dataset performance is better. The typical approach of pre-processing the images with Principal Component Analysis is unsatisfying here for two reasons. Firstly, relying on appropriate training data restricts the general applicability of the system to familiar environments with available training data, or only offline operation. Secondly, from a biological perspective, the PCA pre-processing route is implausible and it is likely other processes come into play. One such possible solution is the use of visual salience; only processing parts of the scene that are likely to be salient. Salience models have been shown to improve navigation performance using conventional SLAM algorithms (Michael Milford, 2014).

While this network was a rate-coded one, the ultimate goal would be to implement the system in a spiking network in order to then access the range of spiking computational hardware currently available and in development.

# 4 Summary

We have presented current ongoing research and results attacking two aspects of the novel sensing problem; the development or adaptation of existing algorithms to successfully use event camera sensory output for useful robotic tasks like navigation, and the development of efficient neural computing architectures to enable these algorithms to eventually be deployed efficiently on neuromorphic hardware. The research is ongoing but promising. Firstly, it appears that "low fidelity" visual place recognitifon techniques like SeqSLAM will readily adapt to being used with event-based cameras. Secondly, it seems likely that algorithms like SeqSLAM can be implemented in relatively lightweight neural models with modest numbers of units. In future we will continue to develop these approaches with the aim of creating a high performance SLAM system that fully takes advantage of neuromorphic computing hardware.